\documentclass[10pt,twocolumn,letterpaper]{article}

\usepackage{wacv}              % To produce the CAMERA-READY version
\usepackage{graphicx}
\usepackage{amsmath}
\usepackage{amssymb}
\usepackage{booktabs}
\usepackage{algorithmic}
\usepackage{algorithm}
\usepackage{textgreek}
\usepackage[T1]{fontenc}

\usepackage[pagebackref,breaklinks,colorlinks]{hyperref}

\usepackage[capitalize]{cleveref}
\crefname{section}{Sec.}{Secs.}
\Crefname{section}{Section}{Sections}
\Crefname{table}{Table}{Tables}
\crefname{table}{Tab.}{Tabs.}

\def\wacvPaperID{2381} % *** Enter the WACV Paper ID here
\def\confName{WACV}
\def\confYear{2025}

\begin{document}

\title{Generating visual explanations from deep networks
using implicit neural representations}

% \author{Michal Byra, Henrik Skibbe\\
% Polish Academy of Sciences\\
% Institution address\\
% {\tt\small mbyra@ippt.pan.pl}}

\author{Michal Byra\textsuperscript{1,2,3}, Henrik Skibbe\textsuperscript{2} \\
\textsuperscript{1}Institute of Fundamental Technological Research, Polish Academy of Sciences, Poland \\
\textsuperscript{2}RIKEN Center for Brain Science, Japan \\
\textsuperscript{3}Samsung AI Center Warsaw, Poland  \\
{\tt\small mbyra@ippt.pan.pl, henrik.skibbe@riken.jp}}

% \and
% Henrik Skibbe\\
% {\tt\small henrik.skibbe@riken.jp}
% }
\maketitle

%%%%%%%%% ABSTRACT
\begin{abstract}
Explaining deep learning models in a way that humans can easily understand is essential for responsible artificial intelligence applications. Attribution methods constitute an important area of explainable deep learning. The attribution problem involves finding parts of the network's input that are the most responsible for the model's output. In this work, we demonstrate that implicit neural representations (INRs) constitute a good framework for generating visual explanations. Firstly, we utilize coordinate-based implicit networks to reformulate and extend the extremal perturbations technique and generate attribution masks. Experimental results confirm the usefulness of our method. For instance, by proper conditioning of the implicit network, we obtain attribution masks that are well-behaved with respect to the imposed area constraints. Secondly, we present an iterative INR-based method that can be used to generate multiple non-overlapping attribution masks for the same image. We depict that a deep learning model may associate the image label with both the appearance of the object of interest as well as with areas and textures usually accompanying the object. Our study demonstrates that implicit networks are well-suited for the generation of attribution masks and can provide interesting insights about the performance of deep learning models.
\end{abstract}

%%%%%%%%% BODY TEXT
\section{Introduction}
\label{sec:intro}

Neural networks have achieved remarkable performance in various computer vision problems. However, explaining deep learning models in a way that humans can easily understand is essential for various applications, especially in medical fields \cite{tjoa2020survey}. Despite their excellent performance, deep neural networks struggle with the well-known 'black box' problem, which undermines confidence in the network's predictions. Methods for explainable artificial intelligence have been intensively studied to help better understand the logic behind the network's predictions. Explainable methods have the potential to detect subtle classification errors, enabling the addressing of unexpected and unwanted behaviors of the networks, and consequently help develop more efficient and trustworthy deep learning models.

Attribution methods constitute an important area of explainable deep learning. The attribution problem involves finding parts of the network's input that are the most responsible for the model's output. In studies on convolutional networks, attribution methods commonly compute saliency maps that highlight input image regions important for the output. Saliency maps assign a score related to prediction importance to each pixel of the input image. The most basic attribution approach is based on perturbing input image pixels and determining the effect of that change on the prediction. Clearly outlining the regions important for the prediction, pointing out the desired objects, is vital for increasing confidence in deep learning methods. However, as highlighted in previous studies, perturbation-based methods are associated with several challenges \cite{fong2019understanding}. Firstly, perturbing all possible combinations of image pixels is infeasible from the computational point of view. As a remedy, the importance mapping is commonly treated as an optimization problem, with carefully designed loss functions and constraints ensuring plausible perturbations. For instance, to avoid adversarial effects, the extremal perturbations technique aims to generate perturbations that have a specific smoothness and size \cite{fong2019understanding}. Secondly, attribution methods usually aim to determine small perturbations that have a potentially large impact on the network's prediction. However, as presented in recent studies, multiple independent explanations might exist for a single image, and determining them might provide additional insight about the deep learning model~\cite{shitole2021one,pomme2023h2o}.

Implicit neural representations (INRs) have recently gained attention in computer vision and medical image analysis (see the Related Work section for a list of potential applications). INRs serve as a continuous, nonlinear, and coordinate-based approximation of the target quantity obtained through a multi-layer perceptron (MLP). Due to this flexibility, implicit networks are especially well-suited for representing complex mappings, for instance representing objects of variable geometry \cite{park2019deepsdf}. Moreover, implicit networks can leverage custom objective loss functions for optimization, jointly addressing various tasks such as image reconstruction and inpainting \cite{nam2022neural}. Due to this versatility, implicit networks have been used to address various complex problems, often requiring case-by-case optimization, which would be difficult to tackle using standard optimization algorithms or convolutional networks that demand large volumes of training data  \cite{sitzmann2020implicit}.
 
\begin{figure*}[]
	\begin{center}
		\includegraphics[width=0.8\linewidth]{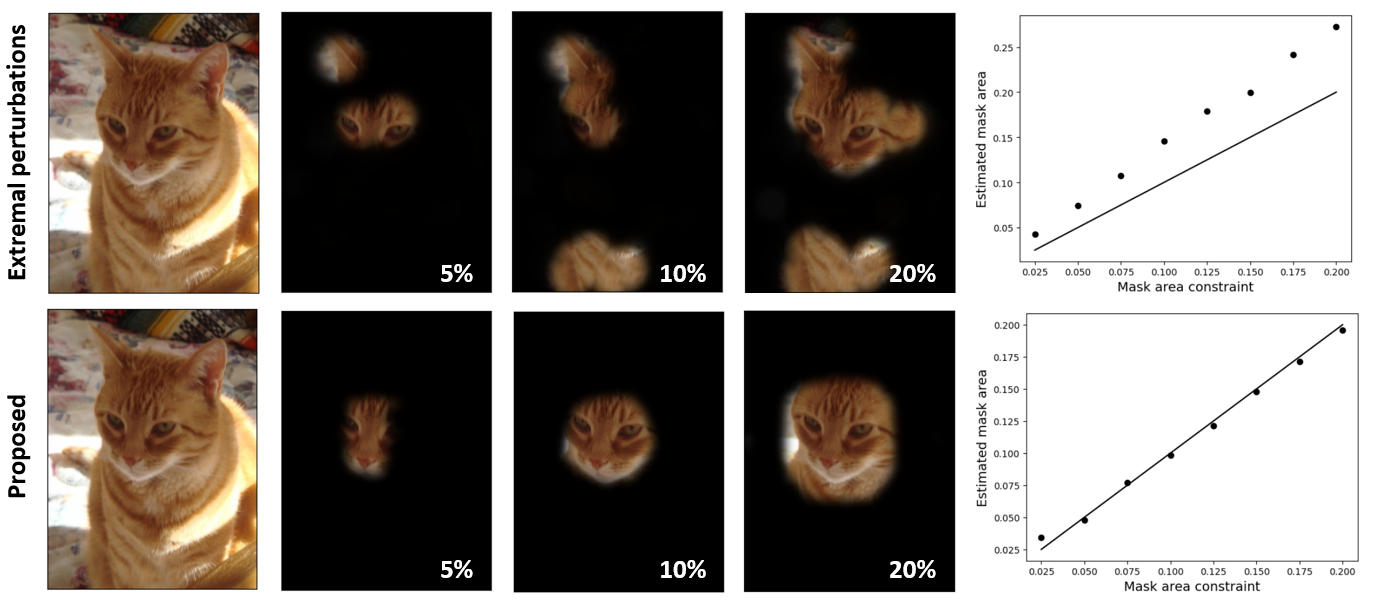}
	\end{center}
	\caption{A comparison between the extremal perturbations technique and the proposed attribution method based on implicit networks, which due to the conditioning mechanism ensures more continuous and well-behaved attribution mask with respect to the mask area constraint. Percentage indicates the area of the attribution mask. }
	\label{fig:1}
\end{figure*}

In this work, we explore the use of implicit networks for explainable deep learning. As far as we know, implicit networks have not been yet used to generate visual explanations. Our main contributions are as follows: 

\begin{itemize}

    \item We demonstrate that INRs constitute a good framework for generating explanations. In comparison to attribution techniques based on standard optimization procedures, implicit networks provide a convenient way to consider non-linear and continuous relationships between the input image pixel coordinates and their importance for the model's prediction. Moreover, implicit networks can be trained using complex custom loss functions, enabling the association of the attribution mapping problem with other computer vision tasks, such as segmentation. 
    
    \item We use INRs to reformulate and extend the extremal perturbations technique \cite{fong2019understanding}. Originally, this technique was introduced to determine attribution masks obtained with an optimization procedure to output a mask that is smooth and covers a specific pre-defined area. However, the optimization procedure has to be repeated for each area constraint to determine mask expansion, which leads to masks that are not continuous with respect to the area constraint. Here, we present that this problem can be mitigated with properly conditioned implicit networks, see Fig. \ref{fig:1}.
    
    \item We present an iterative method based on INRs that can be used to generate multiple explanations for the same input image. This is achieved by utilizing a segmentation-related loss function for the training of the implicit network, which ensures that the newly generated attribution masks do not overlap with the previous explanations. We demonstrate that this approach can provide useful insights about mechanisms guiding network predictions. 
    
\end{itemize}

\section{Related Work}
\label{sec:related}

Explainable artificial intelligence is an intensive area of research in computer vision and medical fields. Below, we discuss the prototypical attribution methods for the selected families of techniques. For a more detailed description of the attribution methods, we refer to one of the recent review papers~\cite{ali2023explainable,saranya2023systematic,szczepankiewicz2023ground}.

\subsection{Attribution methods}

\subsubsection{Perturbation-based methods} 

This family of attribution methods aims to occlude the input image with different types of perturbations and then assess the resulting change in the model's output. For instance, the model-agnostic meaningful perturbations technique optimizes a spatial perturbation mask indicating the image region that maximally affects the output of the model \cite{fong2017interpretable}. The extremal perturbations method extends the meaningful perturbations framework by introducing additional mask smoothing factors and an area loss function~\cite{fong2019understanding}. Leveraging the meaningful perturbation approach, a U-Net-like masking model was trained on ImageNet to generate attribution masks \cite{dabkowski2017real}. While fast at inference, this method requires large volumes of training data, making it infeasible to apply for small datasets, or associating the mask generation problem with other tasks. Moreover, the RISE technique probes the deep learning model with randomly masked versions of the input image to determine image regions important for the predictions and derive a saliency map \cite{Petsiuk2018rise}.

\subsubsection{Activation-based methods} 

These attribution techniques utilize network weights and activations at specific layers to generate saliency maps. For instance, the prototypical Class Activation Map (CAM) technique combines the weights of the last layer with its activations to compute a low-resolution saliency map \cite{zhou2016learning}. Various approaches have been proposed to improve the CAM method, including the popular GradCAM algorithm \cite{selvaraju2017grad}, as well as other extensions, such as AblationCAM \cite{ramaswamy2020ablation}, CAMERAS \cite{jalwana2021cameras}, GradCAM++ \cite{chattopadhay2018grad}, and Score-CAM \cite{wang2020score}, to name a few. Leveraging CAM techniques and perturbation-based approaches, in OptiCAM, the saliency map is optimized on a per-case basis by combining the weights of the model and the activations via a standard numerical procedure \cite{zhang2023opti}.

\subsubsection{Propagation-based methods} 

This family of techniques generates saliency maps based on gradients back-propagated from the selected model's output to the input. In the most basic approach, the back-propagated gradient is considered as a saliency map \cite{simonyan2013deep}. In the guided back-propagation technique, the backward pass for the ReLU activation functions is modified to enhance the gradient flow and, consequently, the saliency map. DeconvNet utilizes deconvolution operations to map prediction-related activations back through the network to the input image space \cite{springenberg2014striving}. More recent methods are based on layer-wise relevance propagation \cite{bach2015pixel} or various approaches to gradient computations, such as in SmoothGrad \cite{smilkov2017smoothgrad}.

\subsection{Implicit neural representations}

Coordinate based implicit networks have proven to be an efficient methods for various problems encountered in computer vision, such as view synthesis \cite{mildenhall2021nerf}, image reconstruction \cite{sitzmann2020implicit}, signal processing \cite{xu2022signal}, shape modelling \cite{yang2021geometry}, image stylization \cite{fan2022unified} and image generation \cite{singh2023polynomial}, to name a few applications. In medical image analysis, neural implicit segmentation functions have been proposed for cardiac segmentation in magnetic resonance imaging \cite{stolt2023nisf}, as well as for image registration \cite{wolterink2022implicit}, image decomposition \cite{byra2023implicit} and vascular modelling \cite{alblas2022going}. In the context of the model interpretability, various studies have been conducted to better understand the mechanics of implicit networks, for example by interpreting implicit networks as Fourier series \cite{benbarka2022seeing} or examining INRs using neural tangent kernel \cite{jacot2018neural,yuce2022structured}. For more applications of INRs, we refer to review papers~\cite{molaei2023implicit,xie2022neural}.

\section{Methods}
\label{sec:prel}

\begin{figure*}[t]
	\begin{center}
		\includegraphics[width=0.9\linewidth]{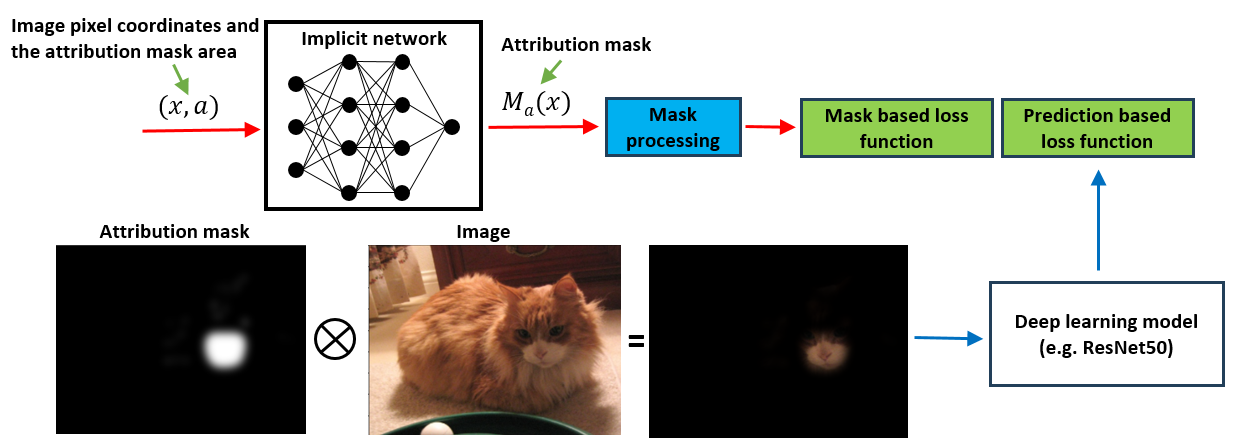}
	\end{center}
	\caption{Scheme illustrating the method proposed in this study. We used a coordinate-based implicit network to compute an attribution mask of size specified by the area parameter. For visualizations, we present blacked-out masks. In implementations, the network processed blurred images, see eq. \ref{eq:image}.}
    \label{fig:2}
\end{figure*}

\subsection{Perturbation analysis with implicit networks}

In this section, we describe our approach to the attribution mask generation with INRs in the context of the extremal perturbations technique. 

Let {\textPhi} stand for the deep learning model we wish to examine with the attribution method. Given a color image $I \in \mathbb{R}^{2}\to\mathbb{R}^{3}$, let {\textPhi}$(I) \in [0,1]$ indicate the post-softmax probability that the input image belongs to the category of interest. In addition, let $\bold{x} \in \mathbb{R}^{2}\to [0,1]^2$ stand for the coordinates of the input image pixels defined on a normalized 2D grid, with $x \in [0,1]^2$ indicating the 2D coordinate of a single pixel. The perturbation analysis deals with finding a mask $M$ which assigns to each input pixel a value $M(x) \in [0,1]$, where $M(x)=1$ indicates that a pixel is important for the prediction and $M(x)=0$ otherwise. Ideally, the mask should point out a coherent part of the image that contributes to the prediction. To assess the importance of the region corresponding to the mask, we modify the input image according to the following equation:

\begin{equation}
    \hat{I}= M \otimes I + (1-M) \otimes I',
    \label{eq:image}
\end{equation}

\noindent where $\otimes$ is the Hadamard product and $I'$ stands for the perturbed image, commonly obtained using either the Gaussian blur perturbation or the fade-to-black zero matrix perturbation.

Following the extremal perturbation method, we consider finding the mask $M$ as an optimization problem, associated with the minimization of the following loss function \cite{fong2019understanding}:

\begin{equation}
    \mathcal{L}_{\text{ext}}(M) = -\text{\textPhi} \big( M \otimes I + (1-M) \otimes I'\big) + \lambda_r R_a(M).
\end{equation}

\noindent The term $R_a(M)$ is a regularization function that enforces the mask area to be equal to $a$, and $\lambda_r$ stands for the weighting parameter. To constrain the area, the values of the mask $M$ are vectorized and sorted in increasing order to form a vector vecsort$_M$~$\in~[0, 1]^{|M|}$, with $|M|$ indicating the number of the mask pixels. To constrain the mask area to $a \in [0,1]$, an auxiliary vector $r_a \in [0, 1]^{|M|}$ is introduced, which consists of $(1-a)|M|$ zeros followed by $a|M|$ ones. Then the regularization function is expressed in the following way:

\begin{equation}
    R_a(M) = \frac{1}{|M|} \sum_{i} \left( \text{vecsort}_M(i) - r_a(i) \right)^2.
\end{equation}

In this work, we utilize coordinate-based implicit networks to represent the attribution masks. The network has the following general architecture: 

\begin{equation}
f_l(x,c)=
    \begin{cases}
       \text{FE}([x, a]), & l=0 \\    
      \rho \left( W^{(l)} f_{l-1}(x,c) + b^{(l)}  \right), & l \in \{1,...,L-1\}  \\
        \sigma \left( W^{(l)} f_{l-1}(x,c) + b^{(l)} \right), & l=L  \\
    \end{cases} 
\label{eq:eq1}
\end{equation}

\noindent where $x$ and $a$ indicate the input 2D coordinates and a vector of area value parameters used to condition the network. $\rho$ and $\sigma$ stand for the ReLU and sigmoid activation functions, respectively. $W^{(l)}$ and $b^{(l)}$ correspond to the weight and bias of the $l$-th layer. FE is the Fourier encoding utilized to compensate for the frequency bias resulting from the utilization of the ReLU activation functions \cite{tancik2020fourier}. We found that inputting the area condition parameters together with the coordinates worked well in our experiments. This approach directly relates the coordinates with the condition parameters and ensures a certain level of smoothness in the output function. 

To obtain a smooth attribution mask $M$ having area parameter of $a$, we imitate the approach from the original study and process the output of the implicit network with a filter based on the radial basis function, $M_a=\text{Filter}(f_L(x,a))$ \cite{fong2019understanding}. The resulting mask is a function of the area parameter. Next, we search for the smallest mask area according to the following formula:

\begin{equation}
a^{*} = \min \left\{ a : \Phi\big( M_a \otimes I + (1-M_a) \otimes I'\big) \geq \Phi_0 \right\},
\end{equation}

\noindent where $\Phi_0$ stands for a post-softmax probability threshold corresponding to a correct prediction. $M_{a^{*}}$ indicates the extremal attribution mask, corresponding to input image area that is sufficient for the network to provide an accurate prediction. Our approach is depicted in Fig. \ref{fig:2}. 

\subsection{Generating multiple explanations}

Multiple explanations might be present for a single image, which may occur in the case of occluded objects or images presenting multiple objects \cite{shitole2021one,pomme2023h2o}. A deep learning model may also assign varying levels of importance to different segments of an object, making predictions based solely on the presence of specific object parts. We propose an iterative algorithm based on implicit networks, which can be used to generate multiple explanations. Given an attribution mask obtained using the perturbation analysis, we tackle the problem of whether we can generate a subsequent attribution mask, which does not overlap with the first one and similarly presents an image region important for the model's prediction. Our approach works in an iterative manner; given a baseline attribution mask $M^b$, we train an implicit network using the following loss function:

\begin{equation}
    \label{eq:mlt}
    \mathcal{L}_{\text{mlt}}(M, M^b) = \mathcal{L}_{\text{ext}}(M) + \lambda_d \mathcal{L}_{\text{dice}}(M, M^b),  
\end{equation}

\noindent where $\mathcal{L}_{\text{dice}}(M, M^b)$ is a soft Dice-based loss function, equal to 0 when $M$ and $M^b$ do not overlap and 1 vice versa \cite{milletari2016v}. $\lambda_d$ stands for the weighting parameter. By utilizing a Dice score-based loss function, we ensure that the new mask is actively pushed to avoid overlapping with the baseline mask $M^b$. Algorithm 1 depicts our iterative approach to attribution mask generation. To determine a subsequent baseline mask, we combine the previously computed masks into a single new mask $M^b$, and then re-train the implicit network using eq. \ref{eq:mlt}. 

\begin{algorithm}[t]
 \caption{Generating multiple attribution masks}
 \begin{algorithmic}[1]
 \renewcommand{\algorithmicrequire}{\textbf{Input:}}
 \renewcommand{\algorithmicreturn}{\textbf{Output:}}
 \REQUIRE Deep learning model $\Phi$, input image $I$, perturbed input image $I'$, initial baseline attribution mask $M^0$, number of attribution masks to generate $N$.
  \FOR {$n \gets 1 $ \TO $N$}
      \STATE $M^b = \sum_{i=0}^{n}M^{i}$ 
      \STATE $M^b = \text{clamp}(M^b, 0, 1)$
      \STATE Train the implicit network from scratch using loss function from eq. \ref{eq:mlt}, utilizing $\Phi$, $I$, $I'$ and $M^b$ 
      \STATE Use the trained implicit network to generate the attribution mask $M_n$
  \ENDFOR
 \RETURN Set of attribution masks $\{M^n\}_{n=0}^{N}$. 
 \end{algorithmic} 
 \label{alg:pretrain_dilation}
\end{algorithm}

\subsection{Evaluations}

\subsubsection{Attribution mask evaluation}

Explainable methods typically serve as a tool for visual inspection of mechanisms governing the prediction generation process. Quantitative evaluation remains challenging as the attribution masks may depend on the performance of the deep learning model and its internal biases. Moreover, multiple visually plausible explanations for a model's prediction may co-exist for a single input image. Common approaches to saliency map evaluation include the pointing game metric and various overlap scores designed to compare the saliency map area with the reference object segmentation. In this work, we used the precision score for the evaluations, which can be expressed with the following equation:

\begin{equation}
\text{Precision} = \frac{|M \cap S|}{|M|},
\end{equation}

\noindent with $M$ and $S$ indicating the attribution mask and the reference segmentation, respectively. By using the precision score, we aim to evaluate if the computed attribution mask is within the reference segmentation mask. Moreover, we introduce a hit rate metric, which we define as a percentage of attribution masks presenting a precision score above 0.5 (at least half of the mask within the reference segmentation), which we believe is more suited for evaluations than the pointing game metric that may produce zero scores both for masks that point out borders of the object and regions slightly outside the reference segmentation. Also, as stated in \cite{fong2019understanding}, a single attribution mask is not suited for the pointing game metric, as the mask commonly does not present a single spatial point corresponding to the maximal saliency score.

In addition, in this study, we trained the implicit network five times for each input image to evaluate the variability of the mask generation procedure with respect to the network weights initialization. Given the five attribution masks, we determined the mean performance scores for each image.

\begin{figure*}[]
	\begin{center}
		\includegraphics[width=0.95\linewidth]{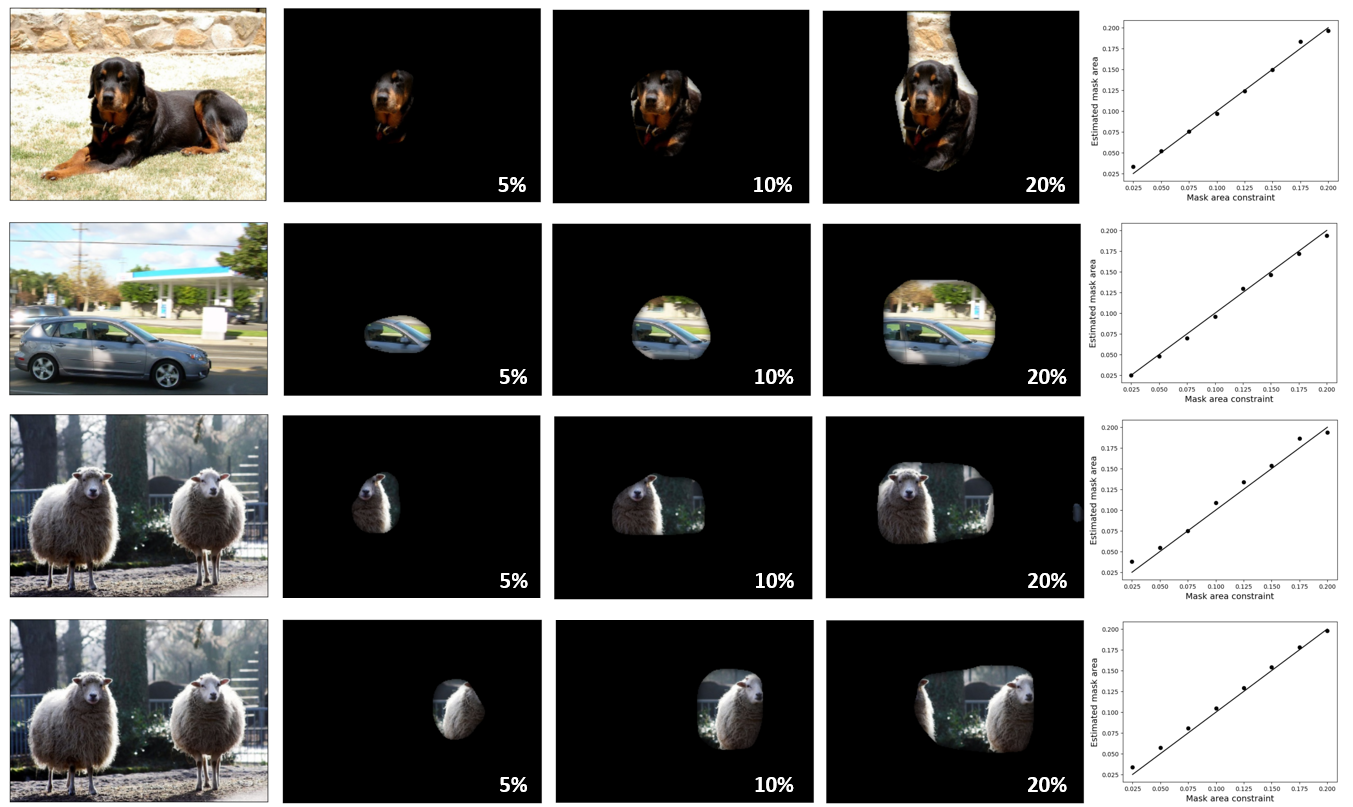}
	\end{center}
	\caption{Illustration of the attribution masks generated with the proposed method. We found that an implicit network could converge to solutions presenting different visual explanations, depending on the network weight initialization. Percentage indicates the area of the attribution mask. }
	\label{fig:4}
\end{figure*}

\subsubsection{Datasets} 

We used the ImageNet-S$_{50}$ validation dataset, which includes detailed semantic segmentations for 752 ImageNet images corresponding to 50 categories \cite{gao2022large}. For the evaluation, we examined the ResNet50 model from the PyTorch model zoo \cite{NEURIPS2019_9015}. Moreover, we also employed the 2007 PASCAL VOC test dataset \cite{everingham2015pascal}. To assess the attribution techniques, we generated rectangular segmentations based on the reference bounding box annotations. For the evaluation, we utilized the ResNet50 model from the TorchRay library, which was pre-trained to classify 20 PASCAL VOC categories~\cite{fong2019understanding}. Following the TorchRay library, we used 2230 images corresponding to difficult cases. Unless explicitly stated, all visualizations presented in this study were generated for the PASCAL dataset.

\subsection{Implementation details}

The perturbed images were generated with a Gaussian filter.  We used the same MLP architecture for all experiments in this study. Each implicit network included five fully connected hidden layers, each with 256 neurons. In addition, we used the Fourier input mapping with six frequencies and 128 components to encode both the spatial coordinates and the area constraint parameter \cite{tancik2020fourier}. Following the extremal perturbation study, we investigated the area constraint parameter in range of [0.025, 0.2], which was scaled before training to [0, 1] to match the range of the coordinates \cite{fong2019understanding}. Adam optimizer with a learning rate of 0.0001 was used to train the networks on a single NVidia 4090 GPU \cite{kingma2014adam}. Each network was trained for 4000 epochs, with each epoch corresponding to a batch of all pixel coordinates and the scaled area parameter uniformly sampled from [0, 1]. $\lambda_r$ and $\lambda_d$ were set to 1. After the training, to determine the attribution mask for the extremal perturbation technique and the proposed method, we examined area parameters equal to \{0.025, 0.05, 0.1, 0.2\} \cite{fong2019understanding}. These two attribution methods were additionally compared with the GradCAM and RISE techniques \cite{selvaraju2017grad,Petsiuk2018rise}, which correspond to popular activation-based and perturbation-based approaches. To generate binary attribution masks for the saliency maps computed with the GradCAM and RISE techniques, we applied thresholding as in \cite{zhou2016learning}. The saliency maps scaled to range of [0,1] were thresholded with manually selected cut-off values of 0.2 and 0.5 for the ImageNet and PASCAL VOC datasets, respectively. All computations were performed using PyTorch 2.1.2 in Python~\cite{NEURIPS2019_9015}. Implementation of the proposed method is available at github.com/mbyr/INR-EXP.

\section{Results}
\label{sec:results}

\subsection{Perturbations with implicit networks}

Fig. \ref{fig:1} visually compares the proposed INR-based method with the extremal perturbations technique. Due to the conditioning mechanism, our approach determines smoother and more continuous attribution masks with respect to the area constraint parameter. In contrast, the extremal perturbations technique computes each attribution mask from scratch for each area parameter, resulting in less spatially continuous masks as different regions can be selected from run to run. Moreover, we found that our method achieved better monotonic correspondence between the area constraint and the actual calculated mask area. In addition, Fig. \ref{fig:4} presents several more examples illustrating the performance of the proposed method. Here, the images from the last two rows demonstrate that the determined explanations may not be unique, as different network weight initialization may result in plausible but non-overlapping attribution masks.

\begin{figure*}[t]
	\begin{center}
		\includegraphics[width=1\linewidth]{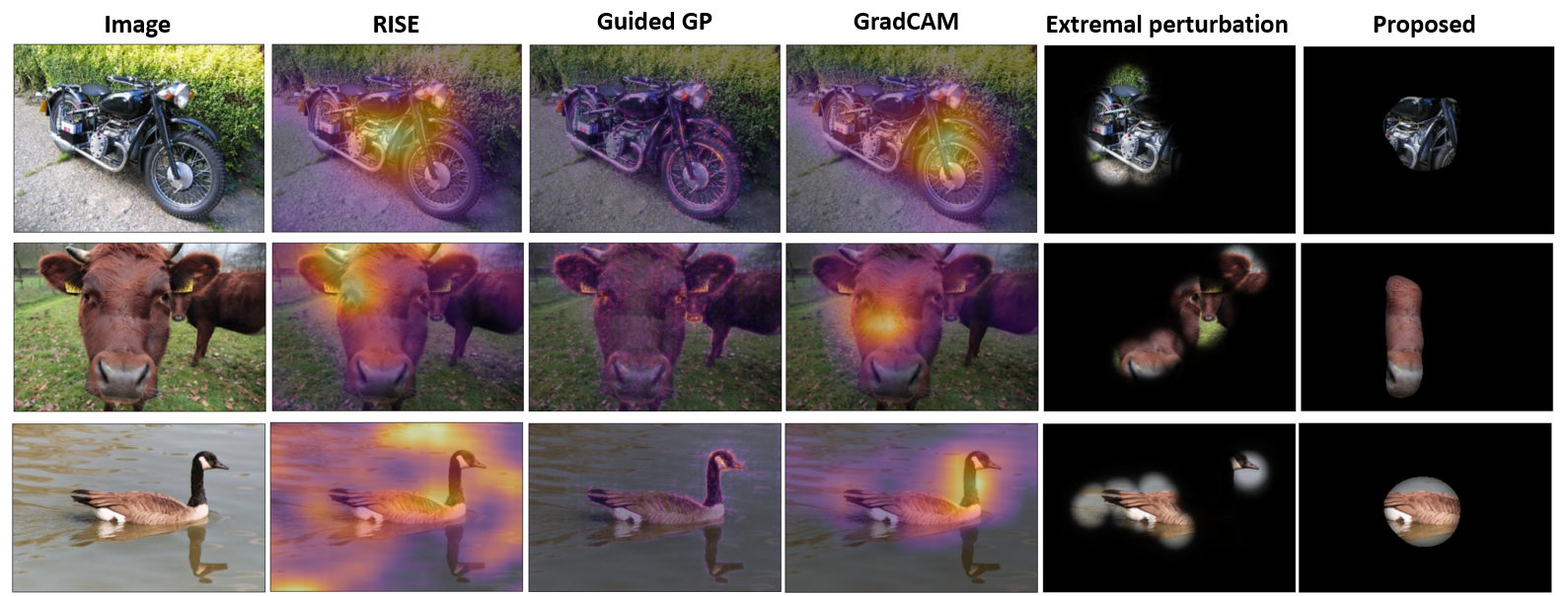}
	\end{center}
	\caption{Qualitative comparison of several attribution methods. }
	\label{fig:3}
\end{figure*}

Qualitative comparison between the proposed method and several other popular attribution techniques is presented in Fig.~\ref{fig:3}. Here, we can notice the diversity between the obtained explanations, associating importance to various parts of the input image. For example, for the results in the second row, obtained for the 'cow' category, saliency maps highlight different parts of the cow's face. Quantitative performance scores are depicted in Table \ref{tab:t1}. For the ImageNet dataset, featuring detailed semantic segmentations, our method outperformed the other techniques, achieving a mean precision score of 0.68. For the VOC dataset, which includes rough bounding box segmentations, we obtained a mean precision score of 0.44 for our method, which was better compared to the RISE and GradCAM techniques but slightly worse than for the extremal perturbation algorithm, 0.46. However, additional statistical analysis based on the t-test ($p<0.05$) presented that there were no significant difference between our method and the extremal perturbation technique on the VOC dataset.  In addition, we evaluated the training time of the implicit network on our GPU, which was equal to approximately three minutes. 

\begin{table}[!t]
  \caption{Precision scores (hit rates) were determined for the investigated methods using the ImageNet and PASCAL VOC datasets. For the ImageNet dataset, which features detailed semantic segmentations, our method outperformed the other techniques. For the VOC dataset, which includes rough bounding box segmentations, our method achieved slightly worse results than the extremal perturbation technique, but the difference was not statistically significant.}
  \label{tab:t1}
  \centering
  \begin{tabular}{|l|c|c|}
    \hline
    Method & ImageNet & VOC \\ \hline
    RISE  & 0.36 (0.28) & 0.39 (0.37) \\ \hline
    GradCAM  & 0.58 (0.60) & 0.38 (0.37) \\ \hline
    Extremal perturbation & 0.63 (0.71) & \textbf{0.46} (0.47) \\ \hline
    Proposed, mean & \textbf{0.68} (0.73) & \textbf{0.44} (0.44) \\ \hline    
  % \hline
  \end{tabular}
\end{table}

\subsection{Multiple explanations}

\begin{figure}[]
	\begin{center}
		\includegraphics[width=1\linewidth]{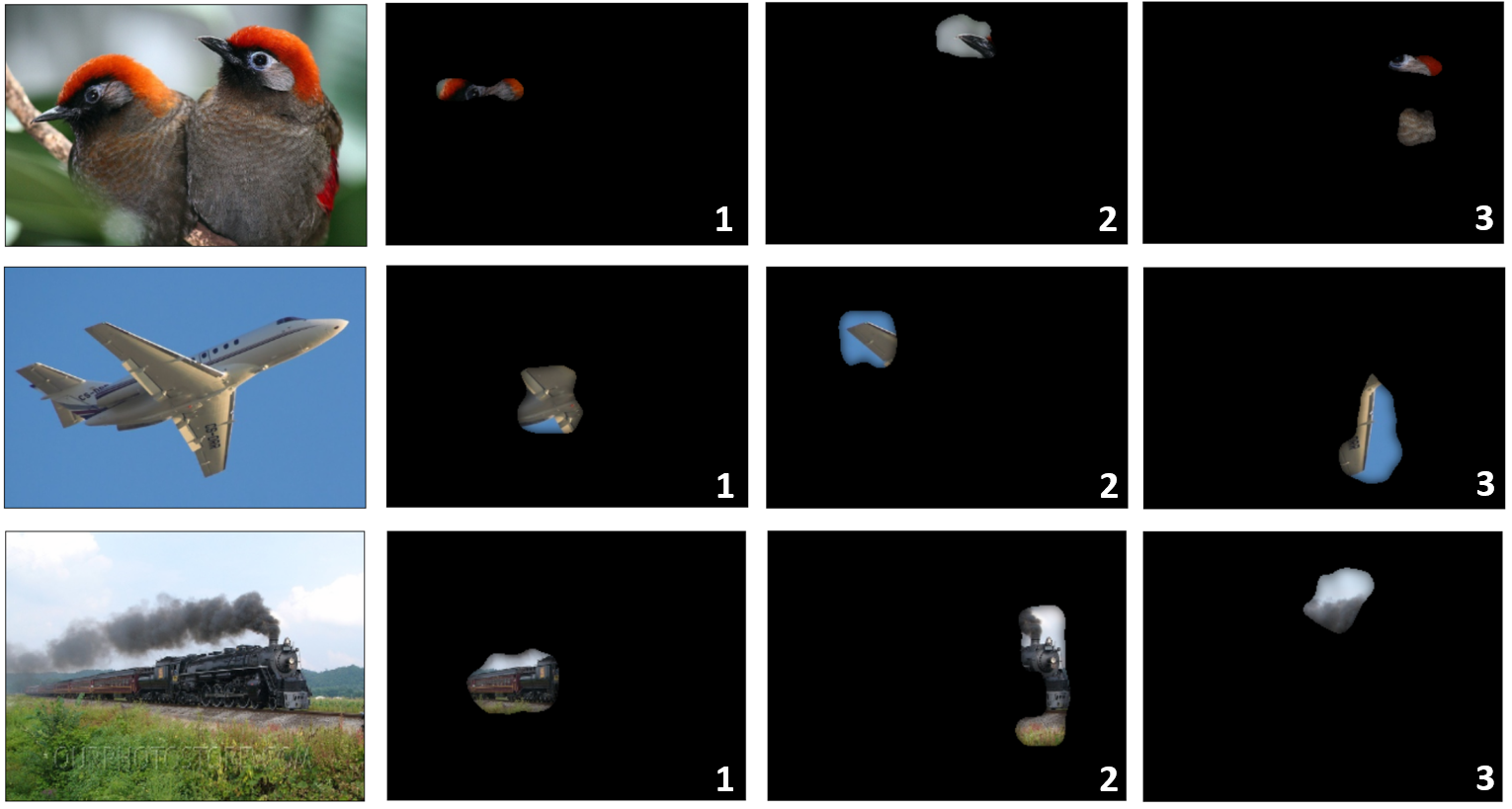}
	\end{center}
	\caption{We used implicit networks to generate multiple non-overlapping attribution masks, separately highlighting different parts of the input image that are important for the prediction.  }
	\label{fig:5}
\end{figure}

\begin{figure}[]
	\begin{center}
		\includegraphics[width=0.9\linewidth]{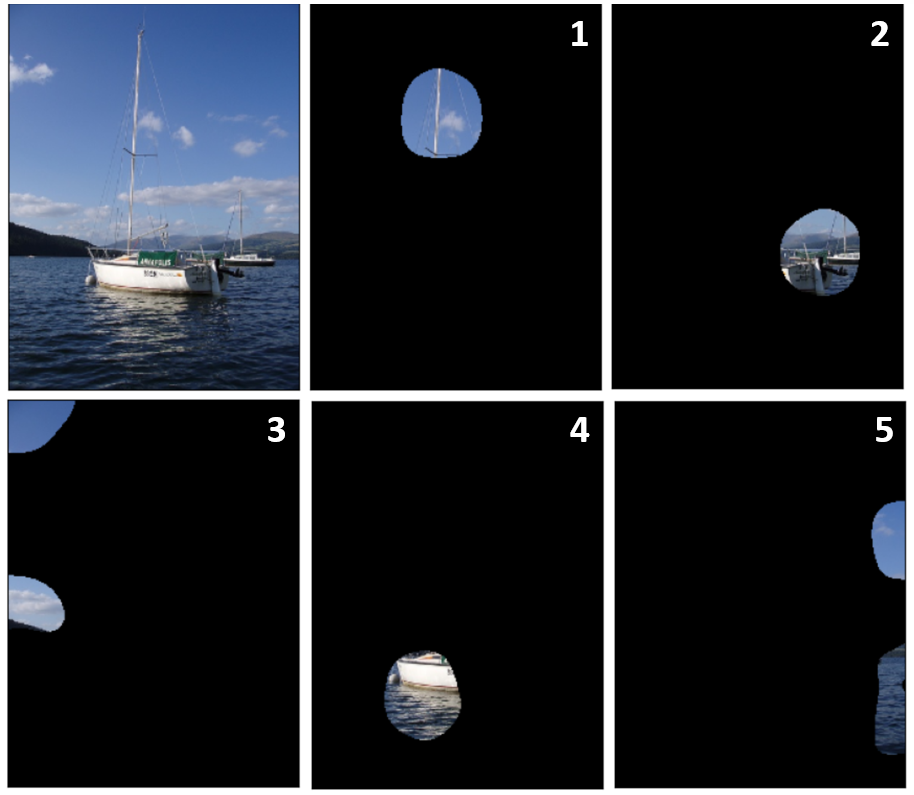}
	\end{center}
	\caption{The proposed technique for iterative explanation generation can be used to provide important insights about the performance of the deep learning models. This example shows that the network associated the 'boat' category not only with the parts of the boat, but also with the sky, clouds and water.}
	\label{fig:6}
\end{figure}

Fig.~\ref{fig:5} illustrates the first three explanations obtained with the proposed iterative approach. The method determined non-overlapping visual explanations related to different parts of the input image. For example, in the second case, to predict the 'plane' category, the network associates importance with different parts of the plane. Fig. \ref{fig:6} shows an image with five explanations, demonstrating an important finding: the network not only associates the 'boat' category with different parts of the boat but also with surroundings commonly accompanying the boat, such as water and blue sky. Notice that for the fifth explanation, the proposed method highlighted both the sky and the water, suggesting that these two regions might be visually compared within the network to provide correct prediction.

We evaluated the consecutive explanations obtained for the ImageNet dataset, with results presented in Table \ref{tab:t2}. The precision score gradually decreased  from 0.68 in the first iteration to 0.30 by the third iteration. This result clearly demonstrates that subsequent explanations tend to be more often associated with the borders of the reference segmentation mask or even with regions outside the mask. However, this was not always the case. Some  attribution masks generated during the second and third iterations showed larger overlaps with the reference mask, evidenced by a maximum precision score of 0.73 (mean score over the maximal precision value of all three subsequent explanations).  

\begin{table}[tb]
  \caption{Performance on the ImageNet dataset. Precision scores (hit rates) were computed for the subsequent attribution masks generated with the proposed iterative procedure (Algorithm 1). Max indicates the performance when selecting the mask with the highest overlap over the three iterations.}
  \label{tab:t2}
  \centering
  \begin{tabular}{|c|c|}
    \hline
    Method & Precision (hit rate) \\ \hline

    Iteration \#1  & 0.68 (0.72) \\ \hline 
    Iteration \#2 &  0.42 (0.42) \\ \hline
    Iteration \#3 &  0.30 (0.24) \\ \hline
    Combination, max &  0.73 (0.77) \\ \hline

  \end{tabular}
\end{table}

\section{Discussion}
\label{sec:discussion}

In this work, we proposed a novel model-agnostic approach to attribution mask generation. Given an input image and a deep learning model, we trained an implicit coordinate-wise network to output the attribution mask highlighting image regions important for the prediction. By using INRs, we extended the extremal perturbations technique. Conditioning the implicit network resulted in attribution masks that are well-behaved with respect to the imposed area constraints. Next, we developed a novel approach for multiple explanation generation. We modified the loss function to train implicit networks to generate non-overlapping attribution masks, pointing out different important parts of the input image. This way, we found that a prediction model may associate the image label with both the appearance of the object of interest as well as with areas and textures usually accompanying the object (e.g., boat vs water). Such findings, demonstrating different operational mechanisms behind the models, are crucial for ensuring robustness in applications.

Utilization of implicit networks offers several advantages for explainable deep learning. Firstly, our study shows that, compared to standard attribution methods, various conditioning mechanisms can be considered in the optimization of the coordinate-wise implicit network, enabling control over the attribution mask generation procedure. Secondly, the training of the implicit network can be performed using various custom loss functions, opening new and interesting directions for the development of explainable methods. Aside from the approach presented in this work, the attribution mask generation procedure could be jointly performed with the input image regression or coordinate-wise image perturbation.

There are several limitations to this work. Firstly, our method explains a 'black-box' deep learning model by training another neural network, which can itself be considered as a 'black-box' model. Therefore, the proposed method might be considered less trustworthy than other attribution techniques. Secondly, training the implicit network requires far more time compared to techniques such as GradCAM, which are based on a single forward/backward pass. However, this limitation could be mitigated by using meta-learning-based weight initialization \cite{sitzmann2020metasdf,tancik2021learned}. Thirdly, while framing the attribution mask generation problem as an optimization task has several advantages, it also presents several challenges, such as the network divergence issue or the requirement to balance loss function components. Although we examined our approach using two datasets, it might be more difficult to converge the network for high-resolution images due to, for instance, GPU memory constraints.

\section{Conclusion}

We believe that our study presents several novel and interesting insights about the explainability of deep learning models. Our work demonstrates that implicit networks are well-suited for the generation of attribution masks. We devised an algorithm that can be used to provide multiple visual explanations, improving the understanding of a network's performance. In the future, we plan to examine different conditioning mechanisms. For example, it would be interesting to utilize implicit networks to associate the attribution mask generation process with other tasks, such as image decomposition or coordinate-wise image perturbation.

\section*{Acknowledgments}

% The authors do not have any conflicts of interest to disclosure.

\noindent This work was partially supported by the programs for Brain Mapping by Integrated Neurotechnologies for Disease Studies (Brain/MINDS) and Multidisciplinary Frontier Brain and Neuroscience Discoveries (Brain/MINDS 2.0) from the Japan Agency for Medical Research and Development AMED (JP15dm0207001,JP23wm0625001).

% *** *** *** *** *** *** *** *** *** *** *** *** *** *** *** *** *** *** *** *** *** *** *** *** *** *** *** *** *** *** *** *** *** *** *** *** ***

%%%%%%%%% REFERENCES
{\small
\bibliographystyle{ieee_fullname}
\bibliography{egbib}
}

\end{document}